\documentclass[letterpaper, 10 pt, conference]{ieeeconf}  

\IEEEoverridecommandlockouts                              

\usepackage{url}
\usepackage{amsmath,amssymb,amsfonts}

\usepackage{cite}
\usepackage{algorithmic}
\usepackage{graphicx}
\usepackage{booktabs}
\usepackage{multirow}   
\usepackage{siunitx}

\usepackage{xcolor}
\usepackage{paralist}
\usepackage{listings}
\usepackage{todonotes}
\usepackage{orcidlink}
\usepackage{paralist}
\usepackage{stmaryrd}
\usepackage{censor}
\usepackage[capitalize]{cleveref}

\newcommand{\RoboVAST}{\textsc{RoboVAST}}

\crefname{section}{Sect.}{sections}
\Crefname{section}{Sect.}{Sections}
\crefname{lstlisting}{listing}{listings}
\Crefname{lstlisting}{Listing}{Listings}
\crefname{figure}{Fig.}{Figs.}
\Crefname{figure}{Fig.}{Figs.}

\lstdefinestyle{mystyle}{
    backgroundcolor=\color{white},   
    commentstyle=\color{codegreen},
    keywordstyle=\color{blue}\bfseries,
    numberstyle=\tiny\color{codegray},
    stringstyle=\color{codepurple},
    basicstyle=\ttfamily\footnotesize,
    breakatwhitespace=false,         
    breaklines=true,                 
    captionpos=b,                    
    keepspaces=true,                 
    numbers=none,                    
    numbersep=5pt,                  
    showspaces=false,                
    showstringspaces=false,
    showtabs=false,                  
    tabsize=2,
    classoffset=0,
    morekeywords={FloorplanGeneration, PathVariationRandom, ObstacleVariation},
    classoffset=1,
    morekeywords={configuration,variations,name,floorplans,path\_length,seed,num\_paths,num\_goal\_poses,obstacle\_configs,amount\_per\_m, max\_distance,args,count},
    keywordstyle=\color{blue}
}

\makeatletter
\def\bstctlcite#1{\@bsphack
\@for\@citeb:=#1\do{%
\edef\@citeb{\expandafter\@firstofone\@citeb}%
\if@filesw\immediate\write\@auxout{\string\citat
ion{\@citeb}}\fi}%
\@esphack}
\makeatother

\begin{document}

\bstctlcite{IEEEBSTcontrol} 

\title{\LARGE \bf
\RoboVAST: Automated Scenario-Based Validation of Robots at Scale
}

\author{Frederik Pasch$^{1}\orcidlink{0000-0002-2626-1538}$ \and Samuel Wiest$^{2}\orcidlink{0009-0002-9215-1238}$ \and Argentina Ortega$^{2}\orcidlink{0000-0002-3873-4435}$ \and Nico Hochgeschwender$^{2}\orcidlink{0000-0003-1306-7880}$
}

\maketitle

\begingroup
    \begin{NoHyper}
  \renewcommand{\thefootnote}{}%
  \footnotetext{$^{2}$ University of Bremen, Germany.
    {\tt\footnotesize \{samuel.wiest, argentina.ortega, nico.hochgeschwender\}@uni-bremen.de}}
  \footnotetext{$^{1}$ Karlsruhe University of Applied Sciences, Germany.
    {\tt\footnotesize frederik.pasch@h-ka.de}}
  \footnotetext{This work has partly been supported by 
    the European Union's Horizon 2020 project SOPRANO (Grant No. 101120990).}
    \end{NoHyper}
\endgroup

\thispagestyle{empty}
\pagestyle{empty}

\begin{abstract}
Validation of robotic systems critically depends on the operating conditions under which they are assessed. Scenario selection and variation are often manual, experience-driven, and difficult to scale, which harms reproducibility and weakens validation conclusions.
We propose a scenario-based methodology that models scenarios compositionally and formalizes how these dimensions are varied, instantiated, executed, and interpreted. Building on this, we introduce \RoboVAST, a framework that realizes declarative campaign specifications, plugin-based scenario generation, and scalable containerized execution with integrated result analysis.
We demonstrate the approach with a navigation dataset comprising 5480 scenario configurations and over 100000 execution runs across five indoor maps with varied paths, sensor noise, software parameters, and obstacle settings, totaling more than 1800 hours of simulated operation and 1873 km traveled. Twenty repetitions per configuration allow us to distinguish systematic failures from stochastic anomalies.
\end{abstract}



\section{Introduction}
\noindent
The responsible deployment of robotic systems—particularly mobile robots—requires rigorous \emph{validation} prior to field operation. 
Validation determines whether a system satisfies requirements such as safety or functional correctness, typically framed as a pass/fail decision~\cite{Araujo2023}, but the reliability of these decisions depends on the diversity and scale of tested conditions.
In practice, validation is commonly carried out using \emph{scenario-based testing}, in which robotic system behavior is observed under concrete operating conditions~\cite{Ortega2024}. 
A \emph{scenario} is a series of interactions between a System Under Test (SUT) and its environment, capturing configurations of environmental conditions, tasks, system parameters, and contextual factors under which system behavior is evaluated~\cite{iso29119pt1}.
Empirical studies show that developers deliberately vary such conditions to uncover corner cases and assess safety and reliability, yet scenario selection is often manual, experience-driven, and difficult to reproduce~\cite{Ortega2022}. 
Simulation-based validation has therefore become central to scenario-based testing~\cite{Afzal2020Testing,Afzal2021}, enabling controlled, repeatable exploration of operating conditions that would otherwise be costly, impractical, or hazardous~\cite{Timperley2018}.
If scenario variation and the number of runs are limited, conclusions may appear valid locally but fail to generalize globally. 

\begin{figure}
    \centering
    \includegraphics[width=1.0\linewidth]{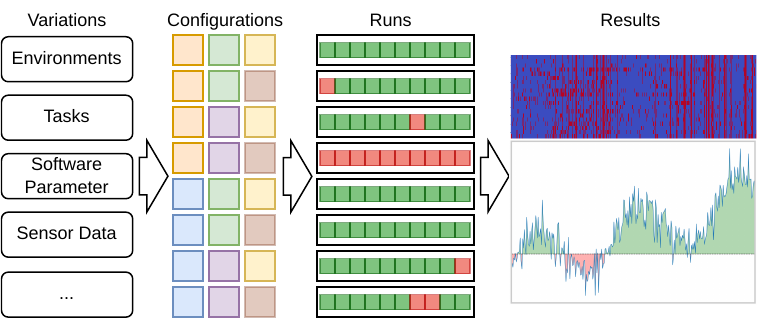}
    \caption{The \RoboVAST\ approach explicitly models scenario variability and enables scalable scenario execution to improve test space exploration.}
    \label{fig:motivation}
    \vspace{-1.5\baselineskip}
\end{figure}

In scenario testing, test coverage is defined as the percentage of executed scenarios out of the total number of scenarios~\cite{iso29119pt1}, yet enumerating all possible scenarios is infeasible due to the combinatorial complexity of system configurations and environmental conditions. 
Scenario specifications often represent only partial descriptions of operating conditions.
This is reflected in two common challenges in testing: \emph{How can meaningful test coverage be defined when the total scenario space is effectively unbounded?} and \emph{how much testing is sufficient to draw reliable conclusions?} Addressing both requires systematic exploration of 
the test space.

However, existing approaches often treat scenarios as monolithic test cases, rather than as compositions of environmental structure, task specification, system configuration, and contextual factors that characterize operating conditions.
As a result, validation results remain tied to individual scenarios and cannot be reliably generalized. 
Without systematic variation across these dimensions and sufficient repetitions, scenario-based testing risks producing locally valid but globally misleading conclusions regarding requirement fulfillment and system performance.
This motivates our central research question: 
\emph{How can scenarios be specified to account for these dimensions, their variations, and their systematic execution?}

This paper makes the following contributions:
\begin{inparaenum}[(i)]
  \item We propose a \emph{scenario-based validation methodology} based on \emph{compositional scenario specifications}, enabling explicit modeling of variability across environmental, task, and system dimensions.
  \item We introduce the \RoboVAST\ testing framework as a declarative \emph{configuration and variation mechanism}, enabling systematic generation and large-scale execution of scenario variations from explicit variability dimensions, as shown in~\cref{fig:motivation}, and supporting coverage and completion criteria in terms of both scenarios and runs.
  \item We provide a large-scale \emph{validation dataset} comprising \num{5480} scenario variants and over \num{100000} total runs with more than 1800 hours of simulated time and 1873 km of traveled distance, demonstrating the scalability of our approach and enabling systematic analysis of variation coverage and run-based confidence.
\end{inparaenum}

\vspace{-0.35\baselineskip}

\section{Related Work}
\noindent Scenario-based testing is a fundamental practice for validating autonomous systems, particularly in autonomous driving systems (ADS) where explicit scenarios enable systematic exploration of operating conditions.
Domain-specific languages such as Scenic and OpenSCENARIO DSL support probabilistic and parametrized scenario specifications~\cite{Fremont2019,dupuis2025}, and safety-focused studies highlight the need for large numbers of executions~\cite{kolb2022}.
In robotics, scenario-based approaches have been applied to mobile robots~\cite{Sotiropoulos2017, Perille2020} 
and drones~\cite{guo2020Sensors}.
Recent work has addressed compositional scenario specification~\cite{Ortega2024} and domain-specific languages for environment generation~\cite{micksei2012,Parra2023}. 
Although parametrized scenarios are well established in autonomous driving, empirical studies show that testing robotic systems still faces substantial obstacles, including limited automation and difficulty in reproducing failures~\cite{Afzal2020Testing}.
Our work aims to bridge this gap by describing how scenario specifications can be parametrized and composed through explicit variability dimensions and automated compositional producers, and demonstrates its implementation in a publicly available tool called \RoboVAST.

Simulation-based testing is essential to modern robotics development~\cite{Afzal2020Testing, Afzal2021}, yet empirical studies reveal persistent challenges including the simulation-to-reality gap, lack of reproducibility, and considerable resource costs~\cite{Afzal2021}. 
While many robotics bugs can be detected in low-fidelity simulation~\cite{Sotiropoulos2017, Timperley2018}, existing test generation approaches~\cite{Khatiri2023, Gambi2019} and benchmarking initiatives~\cite{Perille2020, Durst2022} typically address individual pipeline stages in isolation. 
Recent robot learning advances via visual-language-action models use GPU-accelerated environments like Nvidia Isaac Lab to produce large-scale execution datasets~\cite{Brohan2023, ONeill2025, Jangir2025}, demonstrating the value of scalable, systematic data collection pipelines that can exploit massive parallelism.
In parallel, recent work shows that cloud infrastructure can be used to elastically scale robotics testing to thousands of parallel simulations without upfront hardware investment, amortizing costs across campaigns by paying only for used resources~\cite{zhang2024case}, yet existing solutions such as AWS RoboMaker remain predominantly commercial and closed-source~\cite{Prabha2023}.

Our work extends these efforts by supporting the full testing process: scenario specification, configuration, scalable execution, and evaluation, while introducing explicit variation points to systematically explore the test space.

\section{Approach}
\label{sec:methodology}

\noindent 
To identify a scenario, scenario-based test design techniques often require an underlying model that describes how these interactions take place and/or what actions are performed by the SUT as a result of triggers from other systems~\cite{iso29119pt1}.
However, because interactions are not discrete in robotics, and not all interactions are known in advance and some can only be discovered at runtime (e.g., due to noise, non-determinism), our approach specifies scenarios in terms of scenario elements -- such as environment, tasks, system configurations -- that can be varied and composed to cause a variety of interactions between the robot, the environment, and other agents or objects (e.g. dynamic obstacles).
Our goal is to systematically validate these interactions by exposing the robot to a wide range of contexts resulting from these compositions.
We refer to the associated planning, monitoring, and completion activities collectively as a \emph{test campaign}.
In the remainder of this section, we formalize the concepts in our approach and discuss their implementation in~\cref{sec:robovast}.

\subsection{Scenario Templates and Variation Choices}

\noindent 
\looseness-1
Let us start by defining the scenario dimensions in the test space that we can vary and compose.
Let ${\mathcal D = \{d_1{:}\tau_1, \dots, d_n{:}\tau_n\}}$ be a set of typed variability dimensions,
where each variability dimension $d_i{:}\tau_i$ has a domain ${\mathrm{Dom}(d_i) \subseteq \llbracket \tau_i \rrbracket}$.
We consider the following dimensions: $d_S{:}\Theta_S$ for the robot or SUT $S$ with configuration space $\Theta_S$, $d_\mathcal E{:}\mathcal E$ for the set of environments $\mathcal E$ in which it can operate, and $d_T{:}T$ for the variations on the set of tasks $T$.
With $d_S$ we account for the fact that robot systems are characterized by a huge amount of variability (e.g., varying sensor and actuator configurations, controller parameters, and software variants)~\cite{Garcia2022}.
A configured system instance is denoted as $S(\theta)$ for $\theta \in \Theta_S$.
Environment variability in $d_{\mathcal E}$ accounts for the distinction between the \emph{operating environment} (sometimes referred to as \emph{world}) $w \in W$ and its \emph{environment representation} $e \in \mathcal E$, i.e., $e$ is a model of $w$.
While the world $w$ denotes the concrete physical environment,
validation occurs on an environment representation $e$ that captures exactly those features of an environment that are explicitly modeled, varied, and controlled.
Possible representations for an environment $e$ include structured descriptions of the environment geometry (e.g., maps or floor plans), static and dynamic entities (e.g., obstacles or agents), contextual parameters (e.g., dynamics, friction, and lighting), and semantic regions relevant for task specification.
Finally, $d_T$ captures task-related parameters (e.g., start poses, task specifications like goal poses or paths to be tracked, number of repetitions, etc.).

To enable the composable specification of scenarios that account for $\mathcal D$,
we use three levels of abstraction commonly used in testing terminology for test cases, as follows:
An abstract scenario $a$ describes the scenario in terms of abstract variables $V$. 
At this level, the specification can be used to establish relations and constraints between scenario elements.
A level down in the abstraction, a logical scenario $b$ also describes the parameter ranges ${\mathcal K \rightarrow \mathrm{Dom}(\mathcal D)}$ that can be assigned to $V$. 
Finally, at the lowest level, a concrete scenario $c$ represents a fully parametrized scenario instance with all possible values assigned by a function $\mathcal P: \mathcal K \rightarrow V$.
Note that we can define different $\{\mathcal K_i \rightarrow \mathrm{Dom}(\mathcal D_i) | \mathcal K_i \subset \mathcal K, \mathcal D_i \subset \mathcal D\}$ to be assigned to $V$.

Thus, we define a composable scenario specification as the tuple
${F = (a, \mathcal K, \Pi, \beta, \varphi, \gamma, \omega)}$,
where $\Pi$ denotes the ordered sequence of producers, $\beta$ denotes the set of random seeds required by $\Pi$, $\gamma$ denotes the execution parameters, such as the number of runs $\phi$, and $\omega$ denotes the set of evaluation functions.

We refer to $\Pi = \langle p_1, \dots, p_n \rangle$ as the ordered sequence of producers.
This realization requires that $\mathcal P$ only produces valid assignments $\mathcal P(\mathcal K) \in \mathcal V (\mathcal D)$, i.e.,
$\mathcal V(\mathcal D)$ represents the valid scenario space. Feasibility may require continuous dimensions to be discretized or sampled into a finite scenario space.
We now define the set of admissible variation choices as 
\begin{multline}
\label{eq:variation-choices}
\mathcal V(\mathcal D)
= \Bigl\{\, v : \mathcal D \rightarrow \prod_i \mathrm{Dom}(d_i) \;\Bigm|\; \\
\forall d_i \in \mathcal D; v(d_i) \in \mathrm{Dom}(d_i)
\ \wedge\ \varphi(v)
\Bigr\}
\end{multline}
where $\varphi$ is a constraint predicate restricting admissible combinations of variability choices.
Thus, $\mathcal V(\mathcal D)$
is a total assignment that fully specifies all variability dimensions, ensuring deterministic instantiation and enabling consistent comparison of execution outcomes across the explored scenario space. 
Constraints $\varphi$ may relate dimensions in $\mathcal D$, 
for example by requiring that the start-goal pair of a mobile robot navigation task ($d_T$) is reachable under the generated obstacle layout ($d_{\mathcal E}$) and compatible with the selected sensor configuration ($d_S$).

\looseness-1
The full test space $Q$ is the product of the dimensions in $\mathcal D$, i.e.,  $Q = d_S \times d_{\mathcal E} \times d_T$.
$|Q|$ is the number of all possible scenarios in the test space. 
Thus, we can define test coverage using the valid scenario space ${\mathcal V(\mathcal D) \subset Q}$ for $(\mathcal K, \mathcal D)$ as $\mathrm{coverage} =\left(\frac{|\mathcal C|}{|\mathcal V(\mathcal D)|}\times 100\right)\%$, where $\mathcal C$ is the set of concrete scenarios $c \in \mathcal C$ composed from $F$ and ${\mathcal C \subset \mathcal V(\mathcal D)}$.

\subsection{Instantiation Through Ordered Producers}
\noindent
Given a composable scenario specification $F$, the concrete executable scenario instances $c \in \mathcal{C}$ are obtained through a structured instantiation process by the function $\mathcal P$.
Each producer is a (possibly stochastic) function ${p_i : (\mathcal K,  \beta) \;\rightarrow \mathcal C}$, 
where $\mathcal \beta$ denotes the set of random seeds or stochastic parameters $\eta$ controlling producer sampling procedures; making randomness explicit ensures reproducible scenario instantiation.
Producers are applied in the fixed total order specified by $\mathcal Pi$, reflecting the dependencies between variability dimensions and ensuring their deterministic transformation and composition.
The mapping is partial to capture constructive feasibility limits of $\mathcal V(\mathcal D)$: even when a variation choice satisfies the declarative constraints $\varphi$, a producer may fail to generate a realizable instance under the previously produced conditions.
For example, an obstacle-generation producer applied after map generation may fail if the requested number or size of obstacles cannot be placed without violating the clearance constraints.
Such failures indicate unrealizable scenario instances and are treated as instantiation failures rather than as execution outcomes.

Instantiation is a deterministic loop over the ordered producers: 
starting from an empty configuration set ${\mathcal C^0=\emptyset}$, 
each producer $p_i$ consumes the set of partial configurations $\mathcal C^{i-1}$, 
and the seed $\eta_i$, then applies its generation or variation function to all the partial configurations in $\mathcal C^{i-1}$ for all the values in $\mathcal K_i$, and returns an updated configuration set $\mathcal C^{i}$.
Formally, for ${F = (a, \mathcal K, \mathcal P, \beta, \varphi, \gamma, \omega)}$ with ${\mathcal Pi = \langle p_1,\dots,p_n\rangle}$, this process is defined by

\begin{equation}
\label{eq:scenario-instantiation}
\begin{aligned}
\mathcal C^0 &= \{\emptyset\},\\
\mathcal C^i &= p_i(\mathcal C^{i-1}, \mathcal K_i, \eta_i)
\quad \text{for } i=1,\dots,n,\\
\mathsf{Inst}(\mathcal K,\beta) &= \mathcal C^n.
\end{aligned}
\end{equation}
where $n = |\mathcal P|$. 
\cref{eq:scenario-instantiation} makes explicit how configuration fragments are accumulated across the producer order $\mathcal Pi$ in a combinatorial manner as a producer $p_i$ generates a new partial configuration for each combination of ${\mathcal C^{i-1} \times \mathcal K^i}$.
Hence, the instantiation process in~\cref{eq:scenario-instantiation} creates all the required test data for all the scenarios in $F$.

\subsection{Scenario Execution and Evaluation}

\noindent
Given an instantiated scenario ${c^{n} \in \mathcal C^n}$, the evaluation proceeds by executing the configured SUT.
Execution is modeled as the relation
${\mathsf{Exec} :
\bigl(S(\theta^n) \times \mathcal E^n \times T^n\bigr)
\;\rightarrow\;
\mathcal R}$
where 
$\mathcal R$ denotes the space of raw execution outputs.
Execution instantiates the environment representation $\mathcal E^n$ as a simulation environment,
configures the SUT $S$ with parameters $\theta^n$,
establishes the required preconditions for $T^n$ (e.g., the start pose),
and produces execution outputs $\rho \in \mathcal R$.

To decouple the execution backend from evaluation,
the composable scenario specification $F$ includes a set of evaluation functions that map the execution outputs to the oracle evaluation ${\omega:\mathcal{R}\rightarrow \mathcal{O}}$.
This mapping is a two-step process.
First, an observation function $\mathsf{Obs} : \mathcal R \rightarrow \mathcal T$ maps the raw runs to execution traces.
Then, the scenario traces are used by an oracle function $\mathcal O$
to evaluate the scenario results. 
The definition and implementation of these functions are beyond the scope of this paper.
In general, an oracle $\mathcal O$ can map traces to pass/fail criteria $\mathcal O: \mathcal T \rightarrow \{\mathsf{true}, \mathsf{false}\}$, 
or to metric functions $m : \mathcal T \rightarrow \mathbb{R}$, among others.

Statistical evidence is obtained by evaluating these interpretations over trace sets for the number of runs $\phi$:
\begin{equation}
\label{eq:statistical-analysis}
\bigcup_{i\in\phi}\mathcal O(\rho_i) \mid \rho_i \in \mathsf{Exec}(S(\theta^n), E^{n}, c^{n})
\end{equation}

\section{RoboVAST}\label{sec:robovast}

\begin{figure*}[ht]
    \centering
    \includegraphics[width=\linewidth]{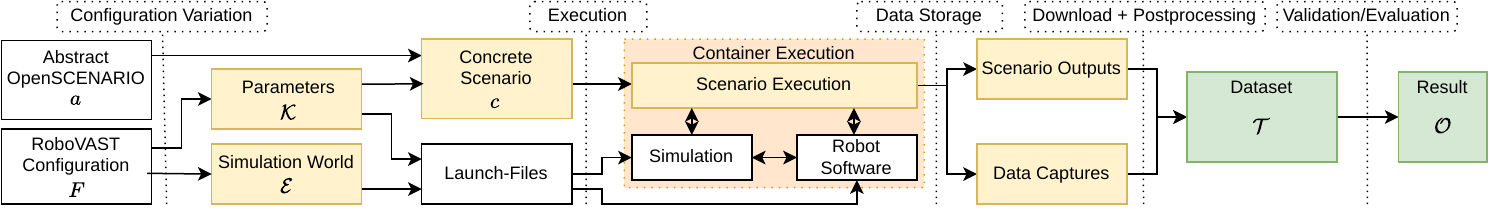}
    \caption{Data flow for a single scenario. The abstract scenario, RoboVAST configuration, launch files, and container images, shown in white, are inputs. The yellow elements describe transient data that is used during the procedure, and the green elements represent the results.}
   \label{fig:dataflow}
\vspace{-1.25\baselineskip}
\end{figure*}

\noindent
To support developers in the validation activities throughout the development life-cycle, we present RoboVAST\footnote{\url{https://purl.org/robovast}}, a framework that implements our approach in~\cref{sec:methodology}.
It can be used for planning (e.g., managing test coverage and/or test completion criteria in terms of $\mathcal{V}(\mathcal D)$ and the number of runs to be executed), 
monitoring and control (e.g., monitoring which test objectives have been met and by which $c \in \mathcal C$ and for which $p_i$), and completion activities (e.g., making available test assets for later use).
\RoboVAST\ is a framework that implements scenario-based validation of robotic systems at scale. It addresses the central challenge of combining expressive scenario specifications with scalable execution to support robust, non-local conclusions about system behavior.  
Concretely, it realizes the methodology defined above through a workflow that combines declarative configuration and variation with Kubernetes-based parallel execution, spanning scenario specification, large-scale execution across many configurations and runs, and evidence interpretation in the form of validation verdicts and performance measurements (cf.~\cref{fig:dataflow}). 
We first outline its core architectural roles and then detail the mechanisms that realize them in \RoboVAST.

\subsection{Conceptual Architecture}

\looseness=-1
\noindent
At the conceptual level, \RoboVAST\ structures testing into three complementary roles: (1) definition of test assets and evidence expectations, (2) declarative configuration and instantiation of scenario families through ordered producers and variability dimensions, and (3) parallel execution of configured system instances across many scenario variants using Kubernetes.
This separation decouples \emph{what} is evaluated (scenario space and criteria) from \emph{how} it is realized (plugins, containers, and orchestration backends). 
\RoboVAST\ therefore provides concrete realizations of the composable scenario specification $F$, 
the ordered producers $\mathcal Pi$, the instantiation function $\mathsf{Inst}(\mathcal K, \beta)$, 
the execution relation $\mathsf{Exec}$, and the observation and interpretation steps given by $\mathsf{Obs}$ and $\mathcal O$.

\subsection{Declarative Specifications}

\noindent
\looseness-1
\RoboVAST\ employs a YAML-based configuration file format as the primary artifact for the specification of $F$.
Concretely, the file declaratively specifies five distinct concerns: (1) \emph{metadata} documenting name, agents and other information related to the test campaign such as description, creators, and licensing information; 
(2) \emph{configuration} defining scenario families and their variability dimensions; 
(3) \emph{execution} specifying containerization, resource allocation, and runtime environment; 
(4) \emph{results processing} declaring postprocessing and publishing pipelines and 
(5) \emph{evaluation} setting analysis-related parameters. 
This separation enables independent evolution of each aspect while maintaining a unified specification of the testing workflow.


The \textit{configuration} section defines one or more scenario families, each with fixed parameters $\mathcal K$ and/or declarative variations that induce systematic parameter sweeps as $\mathcal P$ via plugin-provided variation types.
Independent variation dimensions compose as a Cartesian product, automatically generating all admissible parameter combinations without manual enumeration.
This declarative approach scales to high-dimensional parameter spaces while preserving explicit traceability from configuration to executed instances.
An example \texttt{vast} file is shown in~\cref{lst:vast-file}.
\begin{lstlisting}[caption=A simplified example of a \RoboVAST\ specification., label=lst:vast-file]
configuration:
- name: example
  variations:
  - FloorplanGeneration:
      floorplans:
      - environments/rooms/rooms.fpm
  - PathVariationRandom:
      num_goal_poses: 2
      path_length: 15.0
      seed: 20
  - ObstacleVariation:
      name: static_objects
      obstacle_configs:
      - amount_per_m: [0.4]
        max_distance: [0.0]
        args: width:=0.8, length:=0.8, height:=1.0
      seed: 42
\end{lstlisting}
\vspace{-1.0\baselineskip}

\subsection{Plugin-Based Variation System}

\noindent
\RoboVAST\ implements variability dimensions through an extensible plugin architecture. The plugin layer covers both general-purpose variation producers and domain-specific generators for mobile robot navigation. These plugins realize the producer functions $p_i : (\mathcal K, \beta)  \rightharpoonup \mathcal C$. 
General-purpose plugins support discrete value enumeration over finite lists and stochastic sampling from uniform or Gaussian distributions with configurable bounds. 
Each distribution-based plugin accepts an explicit random seed, ensuring reproducible sampling across campaign executions. For example, we model LiDAR uncertainty by adding zero-mean Gaussian noise with configurable standard deviation to the /scan laserscan topic and by simulating sensor failures via random dropout of individual range measurements, parameterized by a dropout probability.

\begin{figure}
    \vspace{0.2cm}
    \centering
    \includegraphics[width=1.0\linewidth]{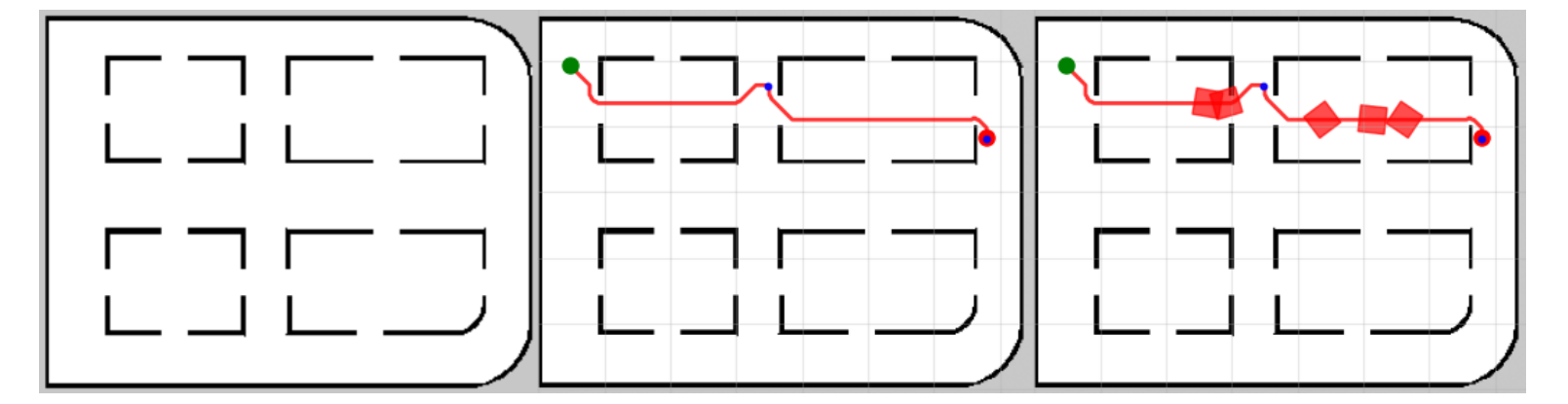}
    \caption{Ordered producer chain $\mathcal Pi$ for one scenario $c \in \mathcal{C}$: from left to right, \textit{FloorplanGeneration} creates the environment representation, \textit{PathGeneration} randomly samples a two-goal task, and \textit{ObstacleVariation} places static obstacles. Together, these steps realize $\mathsf{Inst}(\mathcal K,\beta)$.}
    \label{fig:producers}
\vspace{-1.25\baselineskip}
\end{figure}

Navigation-specific plugins realize these producers for concrete variability dimensions such as maps, tasks, and obstacle settings. \textit{FloorplanVariation} and \textit{FloorplanGeneration} use two user-authored artifacts: a Floorplan DSL definition and a corresponding Variation DSL to generate indoor environments with variation, for example by changing room or doorway dimensions via finite value sets or distribution-based sampling. 
The resulting artifacts, including 2D occupancy grid maps and 3D mesh representations, provide the successive environment states $\mathcal E^{i}$ from the formal instantiation process. 
The \textit{PathVariation} plugins generate start-to-goal paths that are guaranteed to be reachable in the generated map, either by stochastic pose sampling subject to path length and inter-pose distance constraints, or by exhaustive rasterization of the navigable space to enable systematic evaluation across the full environment topology. \textit{ObstacleVariation} places obstacles along generated paths within a configurable proximity envelope. An example is provided in~\cref{fig:producers}, as generated by the vast file presented in~\cref{lst:vast-file}.

The combination of these plugins allows scenario specifications to cover critical corner cases and requirement-relevant conditions, and to validate performance metrics across a family of systematically varied scenarios. 
Moreover, the plugin system enables the extension to other test processes and activities, or other robotic use cases by developers of new producers.

\subsection{Scenario Execution}

\noindent
Configuration generation and runtime parameter resolution realize the instantiation step in \RoboVAST, implementing the formal instantiation function $\mathsf{Inst}(\mathcal K, \eta)$.
During configuration generation, \RoboVAST\ applies the producer plugins to produce the fully resolved OpenSCENARIO parameter values for every configuration variant. As certain producers are depending on others, e.g., the path generation can only succeed if a map was generated, plugins can share their data with others. As the configuration generation is time consuming but deterministic, given identical inputs and seeds, the results are cached internally.
\RoboVAST\ realizes scalable execution through integration with Scenario Execution\cite{Pasch2024} and Kubernetes-based orchestration. This stage implements the execution relation ${\mathsf{Exec}(S(\theta^n), \mathcal E^{n}, T^{n}) \rightarrow \mathcal R}$. Scenario Execution is a middleware-agnostic framework that abstracts simulator-specific details~\cite{Pasch2024}. It parses declarative scenario definitions from OpenSCENARIO DSL into behavior trees, 
which ensure that the necessary preconditions for each scenario are met, monitor and record the test run, and manage the post-execution test activities (e.g., ending a test due to time-outs, or terminating simulation and other processes).
During execution, the instantiated environment representation $\mathcal E^{n}$ is realized as a concrete simulation environment, and configured system instances $S(\theta)$ interact with that environment over time.

To simplify the integration of external tools with \RoboVAST\, a multi-container approach is established. Besides the main container for handling data management, receiving the configuration and sending results, and executing the scenario, a developer can specify any custom container image. Depending on the requirements, only a minimal executable is needed that interacts through UNIX domain sockets with the main container. For each container, resource monitoring is automatically initiated to track metrics such as CPU usage and memory consumption. 
\RoboVAST\ supports both local execution for convenient development and debugging of test campaigns as well as distributed execution on clusters. Furthermore, the containerized architecture allows integration with arbitrary simulators, as long as they can be packaged in a compatible container interface.

To achieve scalability, \RoboVAST\ deploys execution via Kubernetes with Kueue for admission control and resource quota management. The execution workflow comprises four stages: (1) configuration upload to an S3 bucket; (2) Kubernetes Job creation for each configuration-run pair; (3) job queueing and admission by Kueue based on available cluster capacity and job resource requests; and (4) result collection via S3 download and optionally upload to other cloud storages. On startup, each job pulls its configuration files from S3, executes the scenario and finally pushes the results to S3 again. This scalability is essential to execute large-scale testing campaigns with many scenario instances, while the automated orchestration and data handling make managing such extensive campaigns feasible. 

To handle substantial campaigns and subsequent analysis, execution produces a structured results directory with a well-defined layout. Each campaign execution creates a timestamped directory containing: (1) snapshot of all input files enabling campaign reproduction; (2) metadata about execution like execution time, duration, versions, and cluster information; (3) transient files that were created by producers; and (4) per-configuration directories with per-run subdirectories. Each run directory captures JUnit XML test results, resource usage time series, log files, and other files depending on the defined scenario (e.g. ROS bag files). This standardized layout decouples execution results from analysis tools, enabling backend-agnostic postprocessing and evaluation for both validation-oriented verdict computation and performance-oriented metric extraction.

\subsection{Scenario Results}

\noindent
To keep extensive testing campaigns manageable from raw data to evaluation, \RoboVAST\ includes a dedicated results processing stage between execution and result interpretation. This stage bridges the raw run space $\mathcal R$ produced by $\mathsf{Exec}$ and the trace space $\mathcal T$ consumed during interpretation. Its purpose is (1) to transform raw execution artifacts such as ROS bag files, logs, and test outputs into representations that are easier to inspect and analyze such as CSV tables; (2) generate combined metadata descriptions; (3) create or upload publication packages. This separation keeps execution backends independent from analysis logic and allows the same campaign results to be reused across multiple evaluations. 
In practice, this means that oracles can match a particular application, test objective, etc., by implementing the appropriate evaluation functions as postprocessing steps or notebooks, e.g., relying on derived event and oracle inputs, or on derived time series, aggregates, and summary statistics.
For iterative analysis of large campaigns, results-processing artifacts are cached by input hash, allowing derived data to be reused when raw results and configurations remain unchanged. This enables iterative analysis without rerunning the full scenario campaign.

To make extensive campaign results accessible and interpretable, result interpretation in \RoboVAST\ is centered on two approaches: A GUI tool that follows the structure of the output-data and provides developers and researchers the possibility to include custom Jupyter notebooks for visualization.
These notebooks realize the observation and interpretation layer by mapping raw runs or processed artifacts to traces via $\mathsf{Obs} : \mathcal R \rightarrow \mathcal T$, and implementing the oracles $\mathcal O$. Examples include calculating variance from optimal paths or determining the minimum distance to obstacles over the course of a run.
A second option is a Model Context Protocol server, which provides a comprehensive set of tools and resources to large language models to query information. Besides general search and query functions that allow processing of relevant campaign and run data, the navigation plugin provides specific navigation-related tools like map drawing and behavior evaluation.

\section{Case Study Dataset}\label{sec:evaluation_dataset}

\noindent
We evaluate our approach by creating an objective-oriented dataset for mobile robot navigation, reporting results across its scenarios.
Following our methodology, each sub-dataset is defined as a set of scenarios with explicit variation dimensions and constraints, and each is aligned with a specific testing objective.

\setlength{\tabcolsep}{5pt}
\renewcommand{\arraystretch}{0.9}

\begin{table*}[ht]
\vspace{0.2cm}
    \centering
    \caption{Scenario variations for our validation campaign. NTP stands for Nav. to Pose, FP stands for Follow Waypoint}\label{tab:datasets}
    \vspace{-0.5\baselineskip}
    \begin{tabular}{lcccccccccccccccc}
        \toprule
               & & \multicolumn{6}{c}{Paths} &\multicolumn{2}{c}{Task} & Param. & \multicolumn{2}{c}{Sensing Perturbation} & \multicolumn{2}{c}{Obstacles}  \\
         &  & Rasterized & \multicolumn{5}{c}{Random (paths per length)} &  \multicolumn{2}{c}{} & Nav2  & Msg. drop & Noise & Static & Dyn. \\
        Dataset & Configs & Total paths & \qty{5}{\metre} & \qty{10}{\metre} & \qty{15}{\metre} & \qty{20}{\metre} & \qty{30}{\metre} & Type & Loops \\ 
        \midrule
         Performance & 1816 & 100 & 25 & 25 & 25 & 25 & - & NTP & - & x & - & - & - & -\\
         Robustness & 1792 & - & - & 4 & 4 & 4 & - & FP & 0, 1 & - & x & x & - & - \\
         Safety & 1872 & - & - & 3 & - & 3 & 3 & NTP & - & x & x & x & x & x \\
         \bottomrule
    \end{tabular}
\vspace{-1.25\baselineskip}
\end{table*}

\subsection{Structure and Objectives}

\noindent
We organize our test campaign in three \textit{vast} files to demonstrate three distinct test sub-processes.
As such, together with the \textit{osc} file, they describe a specific set of \textit{configurations} that we use to validate a test objective for the same SUT and same test basis -- knowledge used to derive the tests -- i.e., ``test behavior, inputs, environment, pass/fail criteria'', etc.
Our objective is to validate three software qualities: Performance, Robustness and Safety.
Our SUT is a TurtleBot 4 using ROS2 executing navigation tasks.
The datasets use five distinct maps, shown from left to right in \cref{fig:heatmap}. Unlike Nav2’s own system tests, which use a single arbitrary sandbox map, we employ multiple environments to expose failures that depend on environment geometry. The second map is derived from a real-world field-testing campaign in a \qty{7000}{\square\metre} university building~\cite{Parra2023}, providing realistic indoor geometry, while the remaining four are synthetic: a hospital-like lobby, an office-inspired layout from the Floorplan DSL examples, and an artificial corridor network and nested-rooms environment created specifically to stress geometry-induced failure modes.

As a basis we use the default Nav2 configuration and the behaviors of the Nav2 actions \textit{NavToPose} and \textit{FollowWaypoint} for single-goal and multi-goal scenarios, respectively. We use the feedback from these actions as an initial criterion for success and failure, namely if the robot reaches within the default Nav2 goal tolerance of \qty{0.15}{\metre} of the final goal pose.
For each scenario instance, execution traces capture robot state, control behavior, task progress, and failure or safety events.
Our oracles $\mathcal O$ are implemented in .ipython notebooks that consume data from postprocessed rosbags, generating metrics dependent on the dataset. 

Our three datasets are summarized in~\cref{tab:datasets}. 
Variations for the Nav2 parameters refer to the robot radius for the global costmap, with the default value of \qty{0.22}{\metre}; we add an additional variation for \qty{0.175}{\metre}.
The message drop variations have the values \numlist{0.0; 0.05; 0.1; 0.15}, and noise \numlist{0.0; 0.05; 0.1; 0.15} standard deviations.

The performance dataset validates path optimality by quantifying the deviation between executed and planned paths as the root mean squared error (RMSE) between the executed trajectory and an A* reference path sampled along arc length. Navigation tasks use \num{100} rasterized paths plus additional random paths of varying lengths and two global costmap robot-radius values; the environment is inflated by the robot radius before planning to obtain collision-free A* paths.
For each scenario, we derive an RMSE threshold by multiplying the A* path length by the Nav2 local costmap robot radius; runs whose RMSE exceeds this path-length–scaled threshold are classified as performance failures, and the rest as successes.

The robustness dataset aims to validate the robustness of the SUT against sensor degradations by introducing message drop and noise to the laserscan. 
By using the \textit{FollowWaypoint} action, we can vary the number of times a task is repeated, which increases the total execution time and thus the opportunities for navigation failures to accumulate; the default with no repetition is loops$=0$, and we add one variation that repeats the goal sequence once (loops$=1$). 
Over longer runs under degraded sensing, even small errors in perception or control can build up, so repeating the same set of goal poses provides a direct probe of robustness over time. 
Our pass/fail criterion uses a fixed localization-error threshold based on the robot’s size: a run is classified as a robustness failure if its summed localization error (the accumulated difference between the robot’s estimated pose and its ground-truth pose over the run) exceeds \qty{0.3}{\metre}, corresponding to the robot’s diameter; otherwise, it is counted as successful.

The safety dataset validates two safety properties of the SUT: collision and safety-response criteria. It further characterizes safety-related motion behavior by quantifying how close the robot comes to obstacles under safety violations, thereby assessing not only whether a collision occurs but also the margin by which unsafe behavior manifests. We focus on causing safety violations using the \textit{NavToPose} action (with a single goal) while adding variations with static and dynamic obstacles in the SUT's path. For static obstacles, we introduce variations based on density (\numlist{0; 1; 3} per \qty{5}{\metre}) and maximum distance from the A* planned path (\qtylist{0;0.3}{\metre}), with all static obstacles modeled as cubes of size \qty{0.8}{\metre}. For dynamic obstacles, we add a single cube of size \qty{0.5}{\metre} directly on the SUT's path (distance to path \qty{0}{\metre}), spawning it at either \qty{1}{\metre} or \qty{2}{\metre} ahead of the robot along the A* planned path. We use a randomization factor of \num{3} for both types of obstacles, indicating the number of different obstacle position variations that are generated. 
Our pass/fail criterion is derived from the default TurtleBot 4 Nav2 configuration: with a maximum speed of \qty{0.5}{\metre\per\second} and a maximum deceleration of \qty{2.5}{\metre\per\second\squared}, we obtain a stopping distance of \qty{0.05}{\metre}. We then define a minimum safe separation as the sum of the obstacle radius (computed as the maximum distance from the cube center to a corner), the robot radius of \qty{0.15}{\metre}, and this stopping distance; a run is classified as a safety failure if the robot ever comes closer to any obstacle than this minimum distance during execution.

We executed all 3 sub-datasets on an auto-scaling Google Kubernetes Engine cluster with maximum \num{20} C4 worker nodes and a total of \num{1920} CPUs in approximately \qty{7}{\hour}. 
We used \num{20} runs per configuration to obtain statistically meaningful estimates of configuration-specific success rates. Under a simple binomial proportion model with independent runs, \num{20} repetitions bound the standard error of the estimated success probability by approximately \num{0.11} in the worst case (around $p = 0.5$), yielding 95\% confidence intervals of width about $\pm 0.22$. This level of precision, obtained via the normal-approximation confidence interval for binomial proportions, is sufficient to distinguish configurations that fail systematically (e.g., success rates $\le 0.2$) from those that succeed reliably (e.g., $\ge 0.8$).
Per run, two CPUs were allocated to both the simulation and navigation containers, and one CPU was allocated for scenario execution. The gathered data was limited to essential data, omitting large data sources such as laser scans or camera images. In total, approximately \qty{737}{\giga\byte} of data simulating \qty{1800}{\hour} and \qty{1873}{\kilo\metre} of robot movement was captured. 

\subsection{Dataset Analysis}

\begin{figure*}[ht]
    \centering
    \includegraphics[width=1.0\linewidth]{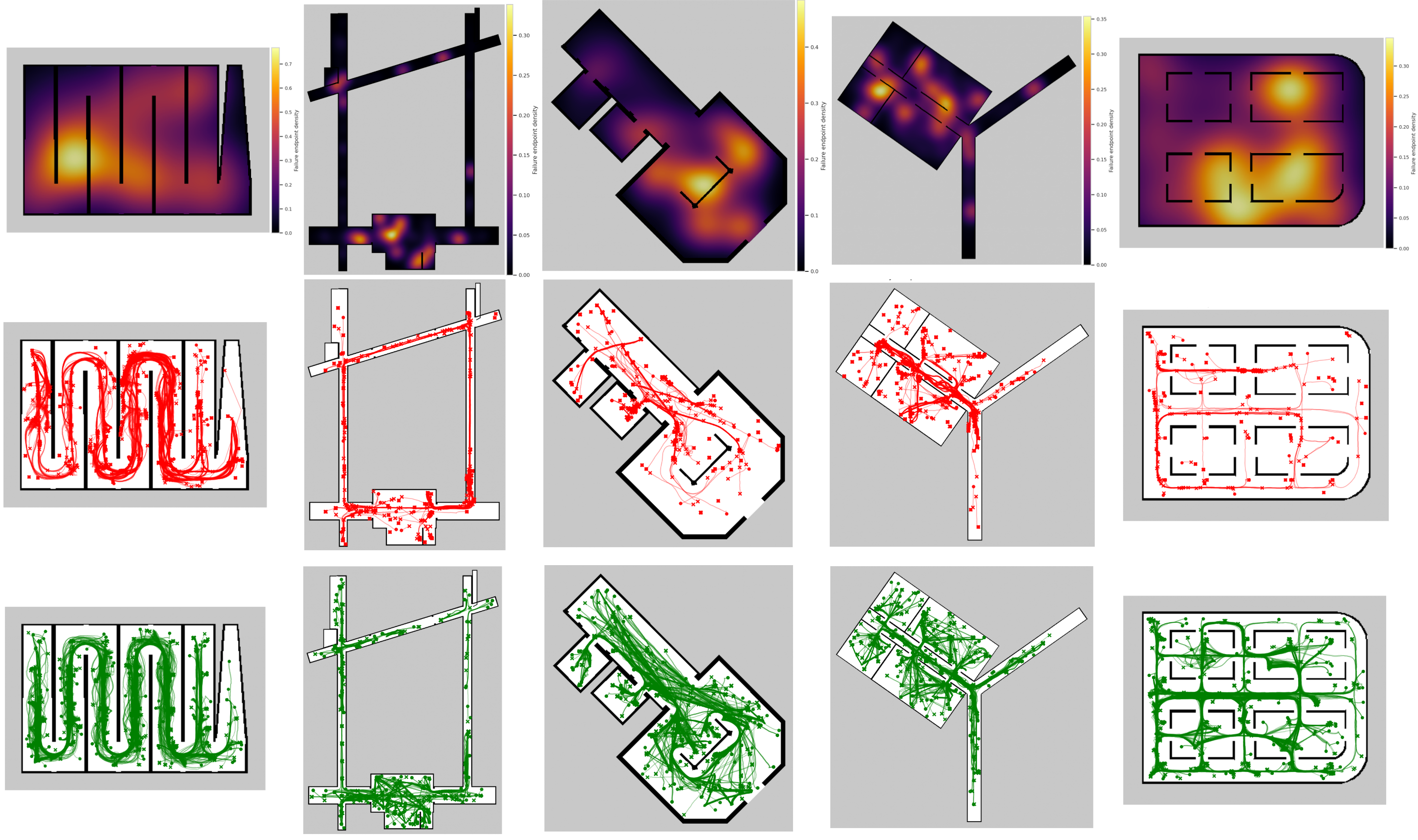}
    \caption{Subset of executions from the dataset, as plotting all executions hinders readability. Top row: final poses of all failed runs across all three sub-datasets. Middle row: \num{1000} failed path trajectories per map, randomly sampled across the three sub-datasets. Bottom row: \num{1000} successful path trajectories per map, randomly sampled across the three sub-datasets. In the middle and bottom rows, circles mark start poses and crosses mark end poses.}
    \label{fig:heatmap}
\vspace{-1.0\baselineskip}
\end{figure*}

\noindent
We analyze the outcomes of the validation campaigns based on dataset-specific success criteria and corresponding success rates, excluding startup failures as they do not relate to the navigation stack. 
The safety dataset yielded \num{15901} successful and \num{10065} failed runs (\num{61.24}\% success); notably, \num{9463} of the failing runs (\num{94.02}\%) belong to scenario variants in which all \num{20} runs fail, indicating that most safety violations are systematic (occurring at high obstacle counts or with dynamic obstacles) rather than random. 
The robustness dataset demonstrates \num{26041} successful and \num{5409} failed runs (\num{82.80}\% success) with \num{32.80}\% of the failures coming from total-failure variants. This indicates that robustness suffered more from random than systematic failures, explainable by the added sensor dropout and noise and long task durations. 
For the performance dataset, results show \num{31714} successes and \num{609} failures (\num{98.12}\% success), with \num{214} failing runs (\num{35.14}\% of all failures) originating from systematically failing variants.

The heatmaps in~\cref{fig:heatmap} visualize the spatial distribution of failures by aggregating final poses of failed runs into failure densities with a blur radius of \qty{0.2}{\metre}. While the success-rate analysis excludes startup failures, the heatmaps include them to show where execution failures occurred spatially.
Regions with high failure density are not simply the most frequently traversed: many successful routes pass through areas with few failures, indicating that failures are tied to specific map features rather than visit frequency.
Failures occur most often at corners, doorways, and near walls, whereas open areas and straight hallways rarely fail; dense spots inside rooms predominantly correspond to startup failures where the robot crashes or terminates before moving.

The complete postprocessed dataset is available at \url{https://doi.org/10.5281/zenodo.19042250}.

\section{Conclusions}

\noindent
\RoboVAST\ is an open-source framework for scalable scenario-based validation of robotic systems. It enables compositional scenario specifications with explicit variation dimensions, supporting large numbers of scenarios for broad test coverage and many runs per scenario to obtain statistical confidence in the resulting safety, robustness, and performance metrics under systematically varied operating conditions.

We demonstrated the approach on a mobile navigation system and released a dataset comprising \num{5480} distinct navigation scenarios and over \num{100000} runs, covering approximately \qty{1800}{\hour} of simulated operation and \qty{1873}{\kilo\metre} of virtual robot motion; the dataset and toolchain are publicly available to support further research on large-scale robotic system validation. Our current evaluation is limited to a single robot platform and navigation stack in quasi-static environments without rich interaction dynamics or multi-robot coordination. Future work includes addressing the sim-to-real gap via comparison of simulated results with real-world tests. We also plan to support additional simulators such as Isaac Sim and O3DE alongside our existing Gazebo integration, extend \RoboVAST\ to other robotic use cases (e.g., manipulation and multi-robot fleets), and explore more systematic combinatorial strategies such as all-pairs testing for covering high-dimensional variation spaces.

\addtolength{\textheight}{-1cm}   


\bibliographystyle{IEEEtran}
\bibliography{references}


\end{document}